\documentclass{article}
\usepackage{algorithmic,algorithm}
\usepackage{appendix} 
\usepackage{amsmath}
\usepackage{amsthm}
\usepackage{amssymb}
\usepackage{amsfonts}
\usepackage{accents}
\usepackage{statex}
\usepackage{booktabs}
\usepackage{bbding}
\usepackage{bm}
\usepackage{bbm}
\usepackage{caption}
\usepackage{tikz}
\usepackage{dsfont}
\usepackage{float}
\usepackage{mathrsfs}
\usepackage{multirow}
\usepackage{multicol}
\usepackage[colorlinks]{hyperref}  
\usepackage{cleveref}
\usepackage{todonotes}
\usepackage[figuresright]{rotating}
\usepackage{makecell}
\usepackage{tikz}
\usepackage{pgf}
\usepackage{subfigure}
\usepackage{setspace}
\usepackage{tabularx}
\usepackage{url}
\usetikzlibrary{arrows, decorations.pathmorphing, backgrounds, positioning, fit, petri, automata}
\usetikzlibrary{trees}
\usetikzlibrary{patterns}

\crefname{equation}{}{}
\crefname{algorithm}{Algorithm}{Algorithms}
\crefname{theorem}{Theorem}{Theorems}
\crefname{lemma}{Lemma}{Lemmas}
\crefname{remark}{Remark}{Remarks}
\crefname{figure}{Figure}{Figures}
\crefname{section}{Section}{Sections}
\crefname{subsection}{Subsection}{Subsections}
\usepackage[top=2.5cm, bottom=2.5cm, left=3cm, right=3cm]{geometry}

\numberwithin{equation}{section}

\newcommand{~}{\,\,\,}

\makeatletter
\def\wideubar{\underaccent{{\cc@style\underline{\mskip10mu}}}}
\def\Wideubar{\underaccent{{\cc@style\underline{\mskip8mu}}}}
\makeatother

\allowdisplaybreaks
\definecolor{Gray}{rgb}{0.5,0.5,0.5}

\begin{document}

\title{Learning More Discriminative Local Descriptors for Few-shot Learning}

\author{
	Qijun Song \thanks{School of Mathematical Sciences, University of Electronic Science and Technology of China, China
		(\href{mailto:songqijun@std.uestc.edu.cn}{songqijun@std.uestc.edu.cn}).}
	\and Siyun Zhou\thanks{School of Mathematical Sciences, University of Electronic Science and Technology of China, China, (\href{mailto:zhousiyun@std.uestc.edu.cn}{zhousiyun@std.uestc.edu.cn}).}
	\and Liwei Xu\thanks{School of Mathematical Sciences, University of Electronic Science and Technology of China, China, (\href{mailto:xul@uestc.edu.cn}{xul@uestc.edu.cn}). The research was supported in part by the National Natural Science Foundation of China (No. 12071060 and 62231016).}}

\date{} 
\maketitle

\begin{abstract}
Few-shot learning for image classification comes up as a hot topic in computer vision, which aims at fast learning from a limited number of labeled images and generalize over the new tasks.
In this paper, motivated by the idea of Fisher Score, we propose a Discriminative Local Descriptors Attention (DLDA) model that adaptively selects the representative local descriptors and does not introduce any additional parameters, while most of the existing local descriptors based methods utilize the neural networks that inevitably involve the tedious parameter tuning.
Moreover, we modify the traditional $k$-NN classification model by adjusting the weights of the $k$ nearest neighbors according to their distances from the query point.
Experiments on four benchmark datasets show that our method not only achieves higher accuracy compared with the state-of-art approaches for few-shot learning, but also possesses lower sensitivity to the choices of $k$.	
\end{abstract}





\maketitle

\section{Introduction}\label{sec1}

Image classification is an important research area in computer vision. Much of the related works rely on collecting and labeling a large amount of data, which is often very difficult and expensive. In addition, such image classification mechanisms are quite different from the human discrimination of images that enables to recognize kinds of targets given only a single image from some certain class \cite{lake2011one}.
These call for not merely an adequate
reduction in the sample size for learning, but also the ability of imitating the intelligent human behavior in image discrimination.
In this context, there is increasing concern about the few-shot learning, which can be generally categorized into three types:
meta-learning based methods \cite{finn2017model,jamal2019task,mishra2017simple,ravi2017optimization,santoro2016meta}, data augmentation methods \cite{NIPS2014_5ca3e9b1,mehrotra2017generative,qi2018low, schwartz2018delta, takahashi2018ricap, vinyals2016matching,zhang2017mixup} and metric-learning based methods \cite{koch2015siamese,li2020more, li2019revisiting,snell2017prototypical,vinyals2016matching,zhang2020deepemd,zheng2023bdla}.
The meta-learning methods aim to provide a model for adjusting the parameters, and thus it can quickly learn new tasks based on the acquired knowledge, while the data augmentation methods pay more attention to the the limited number of the available data in the few-shot learning.
And the metric-based learning methods first map the original images to a high-dimensional semantic space, and then compute the distance between the samples in the support set and the ones in the query set to do the classification tasks.

In this work, we concentrate on the metric-based learning methods.
The literatures \cite{koch2015siamese,snell2017prototypical,vinyals2016matching} all use the image-level features for classification, and the only difference is the metric adopted. 
However, due to the small sample size of each class, the image-level features based methods may not be effective. 
On this account, the Deep Nearest Neighbor Neural Network (DN4) model \cite{li2019revisiting} uses the local descriptors of the image to learn an image-to-class metric and consequently, the model is more capable of catching the discriminative information of the image. 
Recently, 
Zheng et al. \cite{zheng2023bdla} apply the bi-directional local alignment methods to the DN4 model and attain better performance.
Note that in these methods, not all local descriptors are well-valued in the classification task.
For this point, the work \cite{li2020more} uses the Convolutional Neural Networks (CNNs) to generate weights for all local descriptors such that the representative local features can be emphasized.
However, there still exist two major problems to be solved: 
most of this type methods are based on
(i)\ the CNNs for highlighting the representative local descriptors, which introduces extra model parameters, and thus increases the model complexity and the cost for parameter adjustment;
(ii)\ the $k$-NN or variants thereof for classification, which is usually sensitive to $k$.

\begin{figure}[h]%
	\centering
	\includegraphics[width=1.0\textwidth]{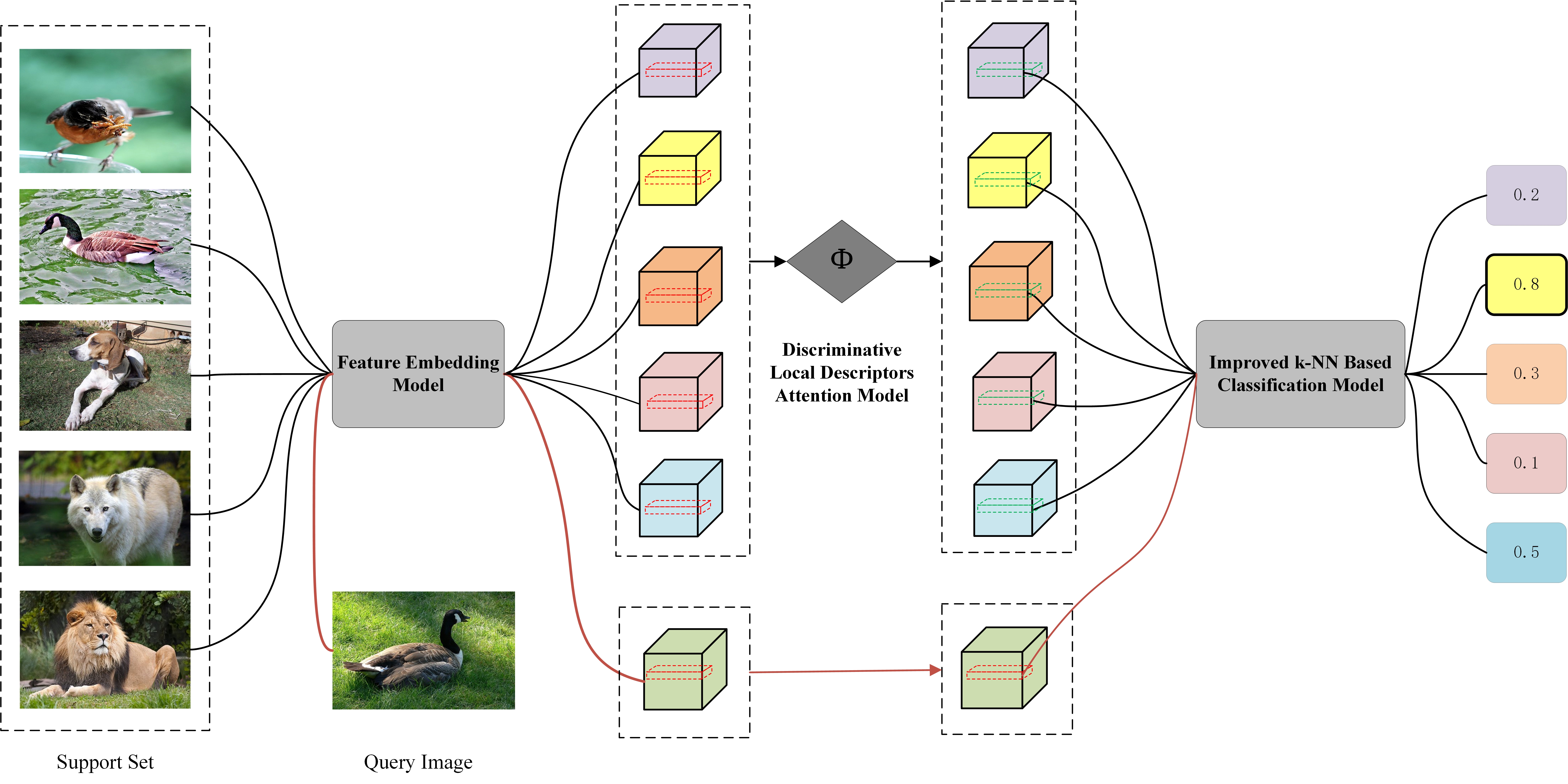}
	\caption{A flow chart of the proposed few-shot learning method for the 5-way 1-shot classification. The learning framework is composed of three models: (i) a feature embedding model implemented on a CNN to extract the local descriptors of images; (ii) a discriminative local descriptors attention model to highlight the representative local features of each class of the images in the support set; (iii) an improved $k$-NN based classification model to measure the similarity between the images in the query set and each class in the support set}\label{fig1}
\end{figure}

To address the above two limitations, in this paper, we propose a Discriminative Local Descriptor Attention (DLDA) model and an improved $k$-NN based classification model. 
As can be seen from Fig. \ref{fig1}, all images are first put into a feature embedding model, and then move to the DLDA model. 
Correspondingly, the DLDA model produces an attention graph for each image, where the discriminative local descriptors are fully valued. This not only stresses the representative local features of each class, but also weakens the effect of noise in the classification. 
The proposed DLDA model is non-parametric, and does not change the size of the feature map. 
In the last stage, the images get into the improved $k$-NN based classification model, where the $k$-NN algorithm finds the nearest $k$ neighbors for each local descriptor in the query set. Then we assign greater weights to those $k$ neighbors that are more concentrated, and thus the impact of $k$ can be reduced.

The rest of this paper is organized as follows. An overview of the related works in few-shot learning is provided in Section \ref{sec2}. The details of the proposed method are presented in Section \ref{sec3}, with extensive numerical experiments following in Section \ref{sec4}. The conclusions of this work is finally given in Section \ref{sec5}.

\section{Related work}\label{sec2}
In the following, we will provide a relatively comprehensive review of existing works on the few-shot learning, which can be classified primarily into three categories as mentioned before.

\noindent\textbf{Meta-learning based methods.} 
Meta-learning \cite{schaul2010metalearning}, also known as  ``learning to learn" \cite{thrun2012learning}, draws on the previous experience to help understand the new tasks.
Finn et al. \cite{finn2017model} propose a Model-Agnostic Meta-Learning (MAML) algorithm to obtain a universal initialization strategy such that the model can converge within just a few number of iterations when faced with a new task. 
However, Jamal et al. \cite{jamal2019task} hold the view that such uniformly-initialized model of the meta-learner can not be well adapted to the new task due to the differences among tasks and in this regard, they design several Task-Agnostic Meta-Learning (TAML) algorithms to prevent the model from over-executing on some certain tasks. 
For the same purpose of getting a better initialization,
Ravi et al. \cite{ravi2017optimization} present a Long Short Term Memory (LSTM) based meta-learner model to optimize another neural network classifier.
Similarly, by taking the advantage of an LSTM-based meta-learner to have access to external memory, Santoro et al. \cite{santoro2016meta} develop a memory-enhanced neural network. 
Although the meta-learning type approaches do spur the development of few-shot learning, it remains a tricky issue in training the models that are of complex memory network structures \cite{mishra2017simple}.

\noindent\textbf{Data augmentation methods.} 
Data augmentation can effectively alleviate the matter of the limited number of samples in few-shot learning 
through constructing new samples based on the old ones. Various geometric transformations can be taken to realize data augmentation, such as flipping \cite{qi2018low}, rotating \cite{vinyals2016matching}, etc. Meanwhile, another technique called Random Image Cropping and Patching (RICAP) \cite{takahashi2018ricap} randomly crops four images and patches them to generate new training images, which owns the advantages similar as the label smoothing.
In \cite{zhang2017mixup}, Zhang et al. propose the mixup method that trains a neural network using the new samples formed by the convex combinations of the samples and their labels. 
By the use of the Delta encoder, the work \cite{schwartz2018delta} is conducted to the extraction of transferable intra-class deformations between the training samples of the same class, and hence the goal of sample synthesis for an unseen class with only few provided samples can be achieved.
Besides, Mehrotra et al. \cite{mehrotra2017generative} apply the generative adversarial networks \cite{NIPS2014_5ca3e9b1} into the few shot learning for data augmentation, and 
provide the generating adversarial residual pairwise networks for the one-shot learning problem.

\noindent\textbf{Metric-learning based methods.} 
Metric-learning approaches are intended for choosing a metric of similarity which computes the distance between the samples in the query set and each class in the support set. They can be roughly divided into two categories: (i) image-level features based methods; (ii) local descriptor features based methods. Among the image-level features based methods,
Siamese neural network \cite{koch2015siamese} extracts the features from two given images and takes the weighted $L_1$-norm to measure the distance between two feature vectors, while the matching network \cite{vinyals2016matching} makes good use of LSTM to enhance the network and improve the learning ability. 
Besides, the prototypical network \cite{snell2017prototypical} is another popular approach based on the image-level features, where the classification is implemented by calculating the Euclidean distance of the class prototypes in the learned embedding space.
Among the local descriptor features based methods, as mentioned earlier,
Li et al. \cite{li2019revisiting} propose a DN4 model to learn an image-to-class measurement. 
Later, Zhang et al. \cite{zhang2020deepemd} investigate the structural distance between local feature representations by using the Earth Movement Distance (EMD) to acquire the image correlation. 
Most recently, a Bi-Directional Local Alignment (BDLA) method is presented in \cite{zheng2023bdla}, which designs a convex combination of the bidirectional distances between the query point and the classes of the support set for classification module.
In addition, to reduce the influence of noises and obtain more representative local descriptors, Li et al. \cite{li2020more} develop a More Attentional Deep Nearest Neighbor Neural Network (MADN4), where the convolutional block attention module is employed for the local descriptor extraction. 

\section{The proposed method}\label{sec3}

\subsection{Problem statement}\label{subsec1}
In the standard few-shot learning, we are given three datasets: 
(i) a training set $\boldsymbol{D} = \left(\boldsymbol{x}_{i}, y_{i}\right)_{i=1}^{N}$ with samples $\boldsymbol{x}_{i}\ (i=1,2,\cdots, N)$ and corresponding labels $y_{i}\ (i=1,2,\cdots, N)$; 
(ii) a support set $\boldsymbol{S} = \left(\boldsymbol{x}_{i}, y_{i}\right)_{i=1}^{MK}$, where $M$ represents the number of classes contained and $K$ is the number of samples in each class; (iii) a query set $\boldsymbol{Q}$, which is comprised of the
samples sharing the same classes as the set $\boldsymbol{S}$ but all unlabeled. And the label space of $\boldsymbol{D}$ and the one of  $\boldsymbol{S}$ are disjoint, thus the sets satisfy $\boldsymbol{Q}\cap \boldsymbol{D}=\boldsymbol{S}\cap \boldsymbol{D}=\varnothing$.
The goal of few-shot learning is to classify the samples in the query set $\boldsymbol{Q}$ according to the support set $\boldsymbol{S}$, and this problem is referred to as ``$M$-way $K$-shot classification". 
Since the number of the samples in $\boldsymbol{S}$ is limited, we use the abundant samples in the training set $\boldsymbol{D}$ to learn the transferable knowledge such that the classification performance of the model on $\boldsymbol{Q}$ can be improved. 

Following the episode training mechanism in \cite{vinyals2016matching}, it is an effective way to make full use of the training set $\boldsymbol{D}$. 
To achieve this, we construct multiple episodes to train our model.
In each episode, we randomly select $M$ classes of samples and $K$ samples belonging to each class from the training set $\boldsymbol{D}$, which form the training support set $\boldsymbol{D}_{S}$,
and take the remaining samples of these $M$ classes in $\boldsymbol{D}$ as the training query set $\boldsymbol{D}_{Q}$.
Once all training episodes have been completed, we use the fully trained model to classify the query set $\boldsymbol{Q}$ according to the support set $\boldsymbol{S}$ in the testing stage.

\subsection{Framework of the proposed method}\label{subsec2}
As can be seen in Fig. \ref{fig1}, our few-shot learning method mainly consists of three parts: a feature embedding model, a discriminative local descriptors attention model, and a classification model. 
Following common practice, the feature embedding model is composed of a CNN, which is used for feature extraction of images in the support set and the query set. 
To emphasize discriminative local descriptors in the image, we introduce an additional attention mechanism model as an intermediate processing, and to improve the classification performance, we incorporate a modified $k$-NN into the last stage, i.e., the classification model, which calculates the similarity scores of the query image with each class of images in the support set, and then predicts the query image to be belonging to the class with the highest score.

\subsubsection{Feature embedding model}\label{subsubsec1}

Given an image $X$, we input it to the feature embedding model, which can be formulated as a mapping $\mathcal{F}_{\theta}$ with the neural network parameter $\theta$.
Then, we get the corresponding output $\mathcal{F}_{\theta}(X) \in \mathbf{R}^{h \times w \times d}$,
where $h$ and $w$ represent the height and width, respectively, and $d$ denotes the number of channels. Note that $\mathcal{F}_{\theta}(X)$ can be written into the following matrix form
\begin{equation}\label{equ02}
\mathcal{F}_{\theta}(X)=\left[\boldsymbol{x}_{1}, \ldots, \boldsymbol{x}_{r}\right] \in \mathbf{R}^{d \times r},
\end{equation}
where $r=hw$, and $\boldsymbol{x}_{i}\in\mathbf{R}^{d}$ is called the $i$-th $(i=1,\cdots,r)$ local descriptor (the red block with dotted lines in Fig. \ref{fig1}) of the feature map of image $X$.
And then we perform a normalization step on each column of $\mathcal{F}_{\theta}(X)$.

\subsubsection{Discriminative local descriptors attention model}\label{subsubsec2}
\begin{figure}[h]%
	\centering
	\includegraphics[width=0.9\linewidth]{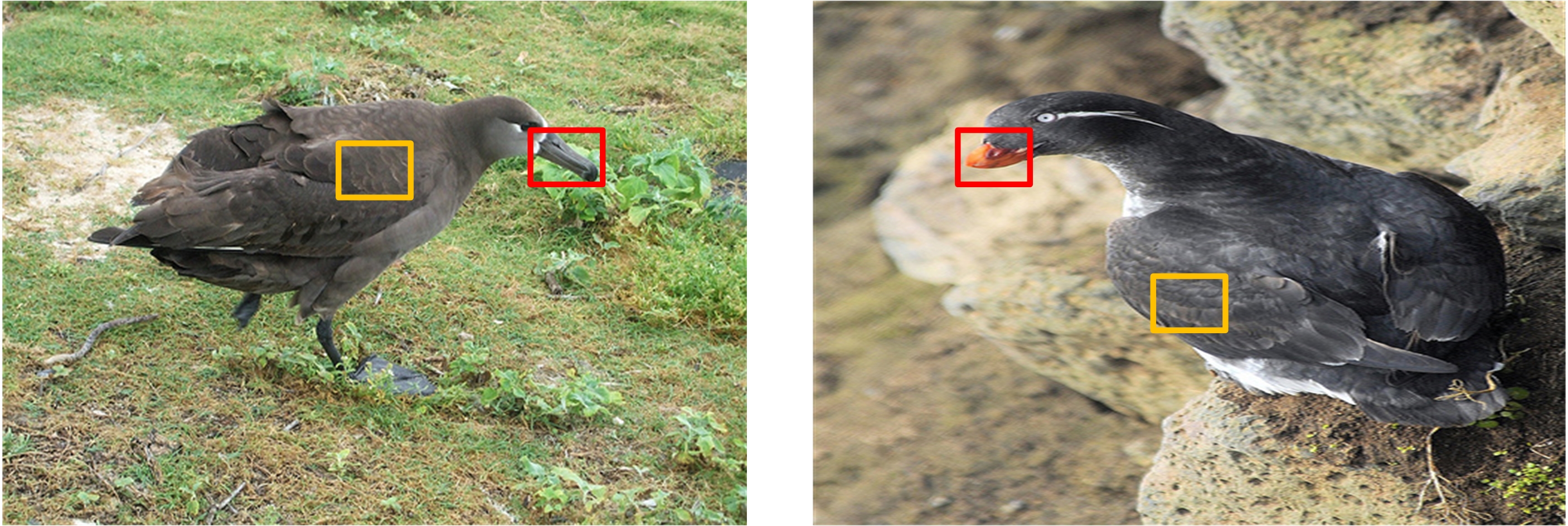}
	\caption{
		Consider the 2-way 1-shot case. The two support images are selected from the CUB-200 dataset, which clearly belong to two different breeds of birds. The local features highlighted in the yellow boxes of the two images are quite similar, and thus the query image is hard to be correctly classified according to this type of local features.
		On the contrary, the local features in the red boxes are more conducive to 
		distinguishing the query image among kinds of images, which naturally play a much more important role in the classification task}
	\label{fig2}
\end{figure}

In some previous studies \cite{li2019revisiting, zhang2020deepemd}, all local descriptors are treated fairly and some underlying representative information of the image may thus be ignored, as illustrated in Fig. \ref{fig2}.
To address this, the work \cite{li2020more} generates the specified weights for local descriptors using CNNs, which however introduces additional parameters that usually requires careful tuning, and may even worsen the overfitting issue in few-shot learning.
At this point, we propose a discriminative local descriptors attention model $\Phi$, which not only underscores the importance of the discriminative local descriptors in the support set, but also avoids the involvement of the extra parameters. 
The proposed DLDA model is inspired by the Fisher Score approach \cite{li2017feature}, where for each local descriptor, the ratio of the intra-class similarity to the inter-class one obtained by $k$-NN is taken as the weight.

Now we take the 5-way 1-shot problem as an example.
Given five images $X_{i}\ (i=1,\cdots,5)$ belonging to five classes that are selected from the support set, 
one has
the output $\mathcal{F}_{\theta}(X_{i})=[\boldsymbol{x}_{i1}, \cdots,\boldsymbol{x}_{ir}] \in \mathbf{R}^{d \times r}$ of the feature embedding model $\mathcal{F}_{\theta}$.
For each local descriptor $\boldsymbol{x}_{ij}\ (j=1,\cdots,r)$, we find its $k$ nearest neighbors, denoted by $\boldsymbol{x}_{ij}^{m}\ (m=1,\cdots,k)$, and then return the corresponding cosine similarities.
Based on that, the intra-class and inter-class similarities are characterized by the following formulas
\begin{align}\label{equ03}
&\text {$intra$-$class$}\left(\boldsymbol{x}_{ij}\right) =  \sum_{m=1}^{k} \cos \left(\boldsymbol{x}_{ij}, \boldsymbol{x}_{ij}^{m}\right),\\
\label{equ04}
&\text {$inter$-$class$}\left(\boldsymbol{x}_{ij}\right) =  \sum_{c \neq i} \sum_{m=1}^{k} \cos \left(\boldsymbol{x}_{ij}, \boldsymbol{x}_{cj}^{m}\right).
\end{align}
Then, the weight $w_{ij}$ of $\boldsymbol{x}_{ij}$ in the DLDA model is defined as
\begin{equation}\label{equ05}
w_{ij} = \frac{\text {$intra$-$class$}\left(\boldsymbol{x}_{ij}\right)}{\text {$inter$-$class$}\left(\boldsymbol{x}_{ij}\right)},
\end{equation}
which finally leads to the weighted feature map of the image $X_{i}$ as
\begin{equation}\label{equ06}
\hat{\mathcal{F}}_{\theta}(X_{i})=[w_{i1}\boldsymbol{x}_{i1}, \cdots,w_{ir}\boldsymbol{x}_{ir}] \triangleq [\hat{\boldsymbol{x}}_{i1}, \cdots,\hat{\boldsymbol{x}}_{ir}] \in \mathbf{R}^{d \times r}.
\end{equation}
For the 5-way 5-shot case where there are five images per class, we first compute an average feature map of the five ones of each class, 
and then follow the same steps as the 5-way 1-shot case. 

In addition, the DLDA model is only performed on the images in the support set, but not on the ones in the query set.
It is also worth noting that the DLDA model is a non-parametric model, which may facilitate the overfitting problem to some extent.

\subsubsection{An improved $k$-NN based classification model}\label{subsubsec3}

The final stage is for classification, and one can choose any appropriate technique to deal with it.
One popular choice is the $k$-NN approach \cite{li2019revisiting,li2020more}, which however implicitly assumes that the $k$ nearest neighbors are of equal importance in the classification decision regardless of their distances from the query point. 
To remedy this, we slightly modify the PNN method \cite{zeng2009pseudo} for the final classification, where different weights are assigned based on the distance from the query point.
The details of the proposed modified $k$-NN based classification model are described as follows.

Given an image $Y$ in the query set, we denote the corresponding output from the feature embedding model as $\mathcal{F}_{\theta}(Y)=\left[\boldsymbol{y}_{1}, \ldots, \boldsymbol{y}_{r}\right] \in \mathbf{R}^{d \times r}$. For each descriptor $\boldsymbol{y}_{s}(s=1,\cdots,r)$, we find its $k$ nearest neighbors $\hat{\boldsymbol{x}}_{ij}^{1},\cdots,\hat{\boldsymbol{x}}_{ij}^{k}$ in class $i$, and compute the corresponding cosine similarity as $\cos(\boldsymbol{y}_{s},\hat{\boldsymbol{x}}_{ij}^{1}),\cdots, \cos(\boldsymbol{y}_{s},\hat{\boldsymbol{x}}_{ij}^{k})$. 
According to the basic idea that a larger cosine similarity means a smaller distance, and therefore a greater weight should be assigned, we then give the formula of the weights for the $k$ nearest neighbors of $\boldsymbol{y}_{s}$ as follows
\begin{equation}\label{equ07}
w_{sn} = \frac{\cos(\boldsymbol{y}_{s},\hat{\boldsymbol{x}}_{ij}^{n})}{\sum_{p=1}^{k} \cos(\boldsymbol{y}_{s},\hat{\boldsymbol{x}}_{ij}^{p})},\quad n=1,\cdots,k,
\end{equation}
where $\cos(\boldsymbol{y}_{s},\hat{\boldsymbol{x}}_{ij}^{n})$ is assumed to be positive for all $1\leq n\leq k$. 
Such assumption is natural and reasonable since the parameter $k$ is often set to a small integer, e.g., $1, 3, 5$.

Finally, the similarity between image $Y$ and class $i$ is defined as
\begin{equation}\label{equ08}
Similarity(Y, \text{class}\ i)=\sum_{s=1}^{r} \sum_{n=1}^{k} w_{sn} \cos(\boldsymbol{y}_{s},\hat{\boldsymbol{x}}_{ij}^{n}),
\end{equation}
which is sum of the $rk$ weighted similarities between $r$ descriptors and their $k$ nearest neighbors.

\subsection{Less sensitivity to $k$ can be expected}\label{subsec3}
As in the traditional $k$-NN, all $k$ neighbors are treated equally, and thus the differences in the distances of these $k$ neighbors from the query point are neglected.
With this in mind, our proposed improved $k$-NN assigns a greater weight to the neighbor that is closer to the query point, or to say, attaches greater importance to the neighbor in the final classification.

\begin{figure}[H]
	\centering
	\begin{minipage}[t]{0.46\textwidth}
		\centering
		\includegraphics[width=5.8cm]{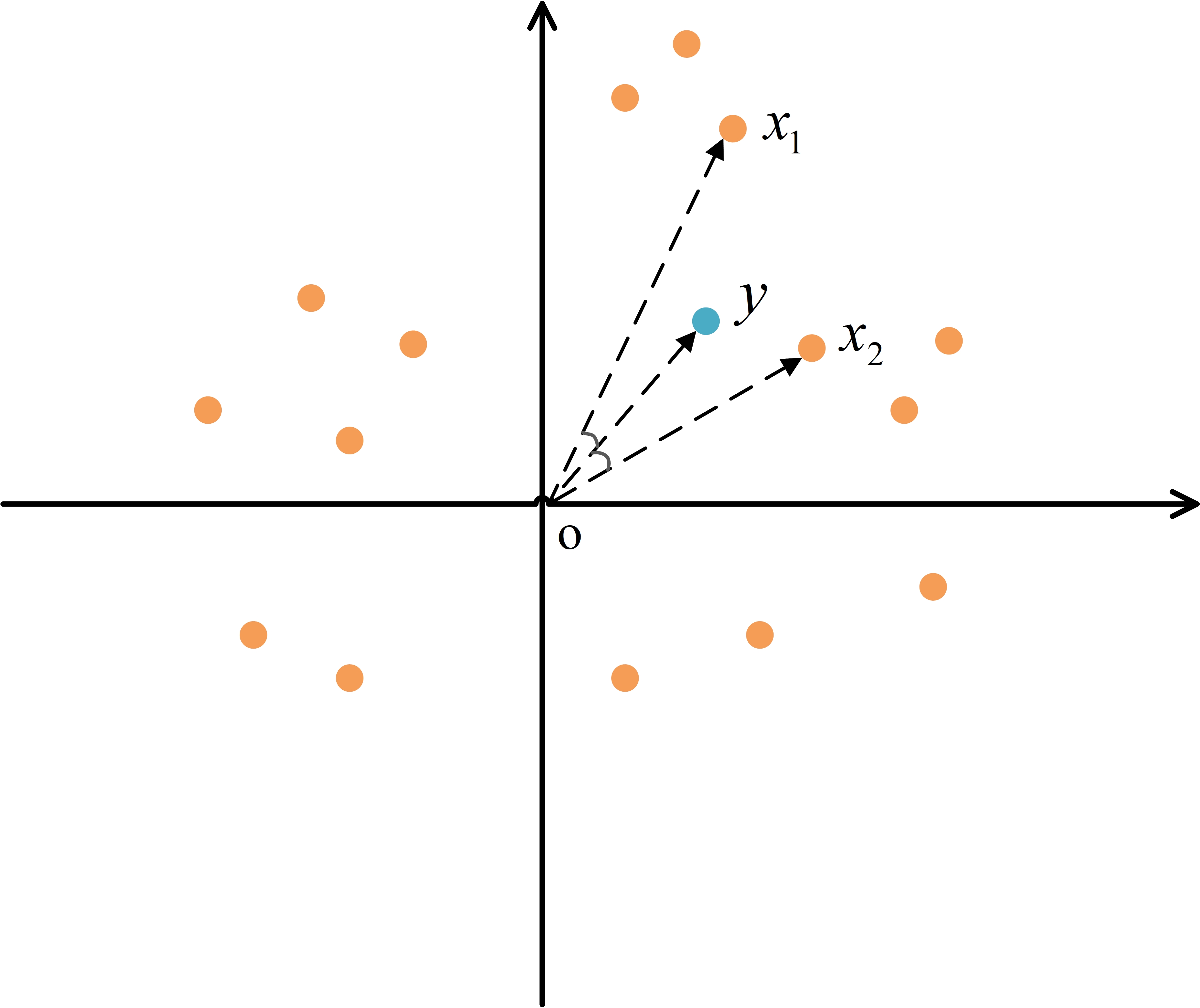}
		\caption{The case of $k=2$: $\cos(\boldsymbol{y},\boldsymbol{x}_{1})$ is slightly larger than $\cos(\boldsymbol{y},\boldsymbol{x}_{2})$}\label{fig3}
	\end{minipage}
	\begin{minipage}[t]{0.46\textwidth}
		\centering
		\includegraphics[width=5.8cm]{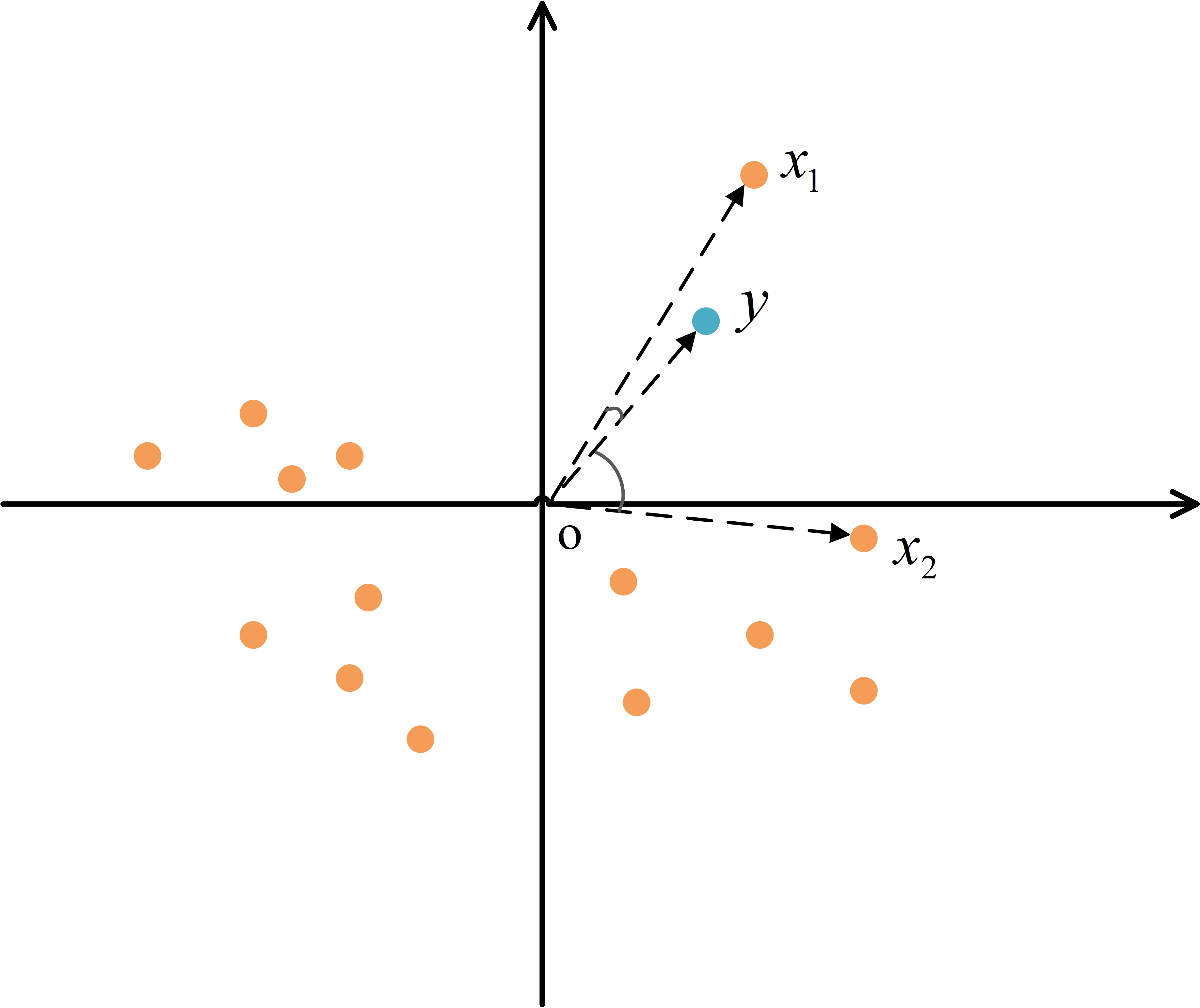}
		\caption{The case of $k=2$: $\cos(\boldsymbol{y},\boldsymbol{x}_{1})$ is much larger than $\cos(\boldsymbol{y},\boldsymbol{x}_{2})$}\label{fig4}
	\end{minipage}
\end{figure}

For illustrative purposes, we consider the simplest $k=2$ case. 
Suppose that the two nearest neighbors of the query point $\boldsymbol{y}$ are $\boldsymbol{x}_{1}$ and $\boldsymbol{x}_{2}$, 
and the relation $\cos(\boldsymbol{y},\boldsymbol{x}_{1}) > \cos(\boldsymbol{y},\boldsymbol{x}_{2})$ holds.
In the traditional $k$-NN, the percentages of the scores of $\boldsymbol{x}_{1}$ and $\boldsymbol{x}_{2}$ can be expressed as
\begin{align}\label{eq:knn}
\boldsymbol{x}_{1}: \frac{\cos(\boldsymbol{y},\boldsymbol{x}_{1})}{\cos(\boldsymbol{y},\boldsymbol{x}_{1}) + \cos(\boldsymbol{y},\boldsymbol{x}_{2})}, \quad\quad  \boldsymbol{x}_{2}: \frac{\cos(\boldsymbol{y},\boldsymbol{x}_{2})}{\cos(\boldsymbol{y},\boldsymbol{x}_{1}) + \cos(\boldsymbol{y},\boldsymbol{x}_{2})},
\end{align}
while in our proposed improved $k$-NN, the percentages can be written as
\begin{align}\label{eq:knn_improve}
\boldsymbol{x}_{1}: \frac{\cos^2(\boldsymbol{y},\boldsymbol{x}_{1})}{\cos^2(\boldsymbol{y},\boldsymbol{x}_{1}) + \cos^2(\boldsymbol{y},\boldsymbol{x}_{2})}, \quad\quad  \boldsymbol{x}_{2}: \frac{\cos^2(\boldsymbol{y},\boldsymbol{x}_{2})}{\cos^2(\boldsymbol{y},\boldsymbol{x}_{1}) + \cos^2(\boldsymbol{y},\boldsymbol{x}_{2})}.
\end{align}
Comparing (\ref{eq:knn}) and (\ref{eq:knn_improve}), we have
\begin{align*}
&\frac{\cos^2(\boldsymbol{y},\boldsymbol{x}_{1})}{\cos^2(\boldsymbol{y},\boldsymbol{x}_{1}) + \cos^2(\boldsymbol{y},\boldsymbol{x}_{2})} - \frac{\cos(\boldsymbol{y},\boldsymbol{x}_{1})}{\cos(\boldsymbol{y},\boldsymbol{x}_{1}) + \cos(\boldsymbol{y},\boldsymbol{x}_{2})} \\
=\ & \frac{\cos^2(\boldsymbol{y},\boldsymbol{x}_{1})[\cos(\boldsymbol{y},\boldsymbol{x}_{1}) + \cos(\boldsymbol{y},\boldsymbol{x}_{2})] - \cos(\boldsymbol{y},\boldsymbol{x}_{1})[\cos^2(\boldsymbol{y},\boldsymbol{x}_{1}) + \cos^2(\boldsymbol{y},\boldsymbol{x}_{2})]}{[\cos^2(\boldsymbol{y},\boldsymbol{x}_{1}) + \cos^2(\boldsymbol{y},\boldsymbol{x}_{2})][\cos(\boldsymbol{y},\boldsymbol{x}_{1}) + \cos(\boldsymbol{y},\boldsymbol{x}_{2})]} \\
=\ & \frac{\cos(\boldsymbol{y},\boldsymbol{x}_{1})\cos(\boldsymbol{y},\boldsymbol{x}_{2})[\cos(\boldsymbol{y},\boldsymbol{x}_{1}) - \cos(\boldsymbol{y},\boldsymbol{x}_{2})]}{[\cos^2(\boldsymbol{y},\boldsymbol{x}_{1}) + \cos^2(\boldsymbol{y},\boldsymbol{x}_{2})][\cos(\boldsymbol{y},\boldsymbol{x}_{1}) + \cos(\boldsymbol{y},\boldsymbol{x}_{2})]} > 0,
\end{align*}
which implies that the proposed improved $k$-NN will enhance the importance of the nearest neighbor and relegate the farthest neighbor to lower importance, and consequently, the final classification will depend more on the nearest neighbor. 
More specifically, we may encounter two possible situations as shown in Fig. \ref{fig3} and Fig. \ref{fig4}. When the cosine similarity between $\boldsymbol{x}_{1}$ and $\boldsymbol{y}$ is close to the one between $\boldsymbol{x}_{2}$ and $\boldsymbol{y}$, the traditional $k$-NN and the improved one are nearly equivalent, but when the cosine similarity between $\boldsymbol{x}_{1}$ and $\boldsymbol{y}$ is  significantly larger than the one between $\boldsymbol{x}_{2}$ and $\boldsymbol{y}$, the improved $k$-NN may have marked effect on the insensitivity to the choice of $k$ as the farther point $\boldsymbol{x}_{2}$ is considered less informative.
Therefore from the discussions above, the improved $k$-NN can behave more stably with respect to different choices of $k\ (k>1)$ than the traditional $k$-NN.

\section{Numerical experiments}\label{sec4}

\subsection{Datasets}\label{subsec4}
\textbf{MiniImageNet} \cite{vinyals2016matching}: 
the dataset is composed of 60,000 images selected from the ImageNet dataset, with 100 classes and 600 images per class. 
Each image is of size $84 \times 84$. 
Following the splitting approach in \cite{ravi2017optimization}, we take 64 classes for training, 16 classes for validation and the remaining 20 classes for testing.

\noindent\textbf{CUB-200} \cite{welinder2010caltech}: 
the dataset covers 200 bird species and the number of images included in each class varies. We take 130 classes for training, 20 classes for validation and the remaining 50 classes for testing.
This dataset is the most widely used benchmark for fine-grained image classification. 

\noindent\textbf{StanfordDogs} \cite{khosla2011novel}:
the dataset consists of 120 breeds of dogs with a total of 20,580 images. We take 70 classes for training, 20 classes for validation and the remaining 30 classes for testing.

\noindent\textbf{StanfordCars} \cite{krause20133d}: 
the dataset contains 16,185 car images of 196 classes, and the classes are mainly derived according to the brand, the model, and the year of manufacture. 
We take 130 classes for training, 17 classes for validation and the remaining 49 classes for testing.

The images in the three fine-grained image classification datasets, i.e., CUB-200, StanfordDogs, and StanfordCars, are resized uniformly to $84 \times 84$, which are consistent with the ones in the MiniImageNet dataset.

\subsection{Experimental settings}\label{subsec5}
\subsubsection{Network architecture}\label{subsubsec4}
For the sake of fairness, we adopt the same network structure of other few-shot learning methods that will be used for comparison in our experiment, e.g., \cite{li2019revisiting,zheng2023bdla}.
To be specific, we use four convolutional blocks to construct the feature embedding model, and each convolutional block contains one convolutional layer with 64 filters of size $3 \times 3$, one batch normalization layer, and one Leaky ReLU layer. 
In addition, a $2 \times 2$ pooling layer is added to the first two convolutional blocks. 
In general, this embedding network is referred to as Conv4.

\subsubsection{Implementation details}\label{subsubsec5}
The experiments are implemented by PyTorch \cite{paszke2019pytorch}. 
Both the 5-way 1-shot and 5-way 5-shot classifications are considered in the experiments. 
At the training stage, we construct 600,000 episodes randomly from the training part of the MiniImageNet dataset and 300,000 episodes from the one of the three fine-grained datasets. 
In each episode, for the 1-shot and 5-shot settings, we choose 1 and 5 support images, 15 and 10 query images from each class, respectively.
Take the 5-way 1-shot setting as an example, we have 5 support images and 75 query images in each episode. 
The Adam optimizer \cite{kingma2014adam} with a cross-entropy loss is used for training.
The learning rate is initialized to $0.001$, and will be reduced by half every 100,000 episodes. 
At the testing stage, we construct 600 episodes randomly from the testing part of each dataset.
The testing process will be repeated 5 times and the average top-1 accuracy along with the 95$\%$ confidence interval will be presented in the results. 


\subsubsection{Baselines}\label{subsubsec6}
To prove the feasibility and superiority of our proposed few-shot learning method, 
for the MiniImageNet dataset, we make comparisons with the following twelve methods: MAML \cite{finn2017model}, TAML \cite{jamal2019task}, Meta-Learner LSTM \cite{ravi2017optimization}, MetaGAN \cite{zhang2018metagan}, GNN \cite{garcia2017few}, TPN-semi \cite{liu2018learning}, Relation Net \cite{sung2018learning}, Matching Net \cite{vinyals2016matching}, Prototypical Net \cite{snell2017prototypical}, DN4 \cite{li2019revisiting} and BDLA \cite{zheng2023bdla}, MADN4 \cite{li2020more}.
And for the three fine-grained datasets, we make comparisons with the following five methods: Matching Net \cite{vinyals2016matching}, Prototypical Net \cite{snell2017prototypical}, DN4 \cite{li2019revisiting}, BDLA \cite{zheng2023bdla} and GNN \cite{garcia2017few}.

\subsection{Results}\label{subsec6}
\subsubsection{Comparison on the MiniImageNet dataset}\label{subsubsec7}
The results on MiniImageNet are given in Table \ref{tab1}. 
Compared with the classical metric-based methods, in the case of 5-way 1-shot, the proposed DLDA model with $k=1$ gains $2.37\%$, $9.25\%$ and $3.39\%$ improvements over Relation Net \cite{sung2018learning}, Matching Net \cite{vinyals2016matching} and Prototypical Net \cite{snell2017prototypical}, respectively. 
And for the more recent DN4, our DLDA model improves the accuracy by $1.57\%$ and $0.74\%$ in the 5-way 1-shot and 5-way 5-shot settings, respectively, which suggests that the DLDA model is more able to emphasize the discriminative local features.
Moreover, the combination of the DLDA model and the improved $k$-NN algorithm further enhances the classification accuracy and outperforms all of the previous methods. Note additionally that though the proposed method has similar results as the ones of MADN4 \cite{li2020more}, our method does not introduce any new parameters, which thus avoids the tedious parameter tuning and may be beneficial to alleviate the problem of overfitting to some degree.

\begin{table}[h]
	\caption{Average accuracy with 95\% confidence intervals on the MiniImageNet dataset}\label{tab1}
	\begin{tabular*}{\textwidth}{@{\extracolsep\fill}lccc}
		\toprule%
		\textbf{Model}  & \textbf{Embedding} & \multicolumn{2}{c}{\textbf{5-Way Accuracy (\%)} } \\
		\cmidrule{3-4}  
		&  & 1-shot  & 5-shot  \\
		\midrule
		MAML\cite{finn2017model} &  Conv4-32 & 48.70 ± 1.84  & 63.11 ± 0.92 \\
		TAML\cite{jamal2019task} & Conv4 & 51.73 ± 1.88 & 66.05 ± 0.85 \\
		Meta-Learner LSTM\cite{ravi2017optimization} & Conv4-32 & 43.44 ± 0.77 & 60.60 ± 0.71 \\
		MetaGAN\cite{zhang2018metagan} & Conv4-32 & 52.71 ± 0.64 & 68.63 ± 0.67 \\
		GNN\cite{garcia2017few} & Conv4-64 & 49.02 ± 0.98 & 63.50 ± 0.84 \\
		TPN-semi\cite{liu2018learning} & Conv4-64 & 52.78 ± 0.27 & 66.42 ± 0.21 \\
		Relation Net\cite{sung2018learning} & Conv4-64  & 50.44 ± 0.82  & 65.32 ± 0.70 \\
		Matching Net\cite{vinyals2016matching} & Conv4-64 & 43.56 ± 0.84  & 55.31 ± 0.73 \\
		Prototypical Net\cite{snell2017prototypical} & Conv4-64 & 49.42 ± 0.78  & 68.20 ± 0.66 \\
		DN4\cite{li2019revisiting} &  Conv4-64 & 51.24 ± 0.74 & 71.02 ± 0.64 \\
		BDLA\cite{zheng2023bdla} &  Conv4-64 & 52.97 ± 0.35 & 71.31 ± 0.68 \\
		MADN4\cite{li2020more} &  Conv4-64 & \textbf{53.20 ± 0.52} &  71.66 ± 0.47 \\
		\cmidrule{1-4}  
		DLDA($k=1$) &  Conv4-64 & 52.81 ± 0.79 & 71.76 ± 0.66\\
		DLDA+Improved $k$-NN($k=3$) &  Conv4-64 & \textbf{53.20 ± 0.82}  & \textbf{71.76 ± 0.47} \\
		\bottomrule
	\end{tabular*}
\end{table}

\subsubsection{Comparison on fine-grained datasets}\label{subsubsec8}
One distinguishing feature of fine-grained datasets is 
the small inter-class variation but the large intra-class variation, which hence makes the classification much more challenging.
As can be seen from Table \ref{tab2}, in the case of 5-way 1-shot, the DLDA improves the accuracy by $4.03\%$, $1.02\%$, $8.28\%$ over DN4 in the Stanford Dogs, Stanford Cars and CUB-200, respectively.
And the one with the improved $k$-NN is ahead of DN4 under both of the 1-shot and 5-shot cases on three fine-grained datasets.
Particularly, on the Standford Dogs dataset, the accuracy is enhanced by $3.67\%$ and $6.65\%$ in the 1-shot and 5-shot classifications, respectively.
Overall, although the DLDA with/without the improved $k$-NN lags behind the BDLA in the 1-shot case on the Standford Cars dataset,
it is safe to say that our method is the winner among all competitors.

\begin{sidewaystable}
	\caption{Average accuracy with 95\% confidence intervals on the fine-grained datasets}\label{tab2}
	\begin{tabular*}{\textwidth}{@{\extracolsep\fill}lccccccc}
		\toprule%
		\textbf{Model}  & \textbf{Embed.} & \multicolumn{6}{c}{\textbf{5-Way Accuracy (\%)} } \\
		\cmidrule{3-8}  
		&  & \multicolumn{2}{c}{\textbf{Stanford Dogs}}  & \multicolumn{2}{c}{\textbf{Stanford Cars}} & \multicolumn{2}{c}{\textbf{CUB-200}} \\
		\cmidrule(r){3-4}  \cmidrule(r){5-6} \cmidrule(r){7-8}
		&  & 1-shot  & 5-shot & 1-shot  & 5-shot & 1-shot  & 5-shot  \\
		\midrule
		Matching Net\cite{vinyals2016matching} & Conv4-64 & 35.80 ± 0.99 & 47.50 ± 1.03  & 34.80 ± 0.98 & 44.70 ± 1.03 & 45.30 ± 1.03  & 59.50 ± 1.01 \\
		Prototypical Net\cite{snell2017prototypical} & Conv4-64 & 37.59 ± 1.00 & 48.19 ± 1.03  & 40.90 ± 1.01  & 52.93 ± 1.03  & 37.36 ± 1.00  & 45.28 ± 1.03 \\
		GNN\cite{garcia2017few} & Conv4-64 & 46.98 ± 0.98  & 62.27 ± 0.95  & 55.85 ± 0.97  & 71.25 ± 0.89  & 51.83 ± 0.98 & 63.69 ± 0.94 \\
		DN4($k=1$)\cite{li2019revisiting} & Conv4-64 & 45.41 ± 0.76 & 63.51 ± 0.62 & 59.84 ± 0.80 & 88.65 ± 0.44 & 46.84 ± 0.81  & 74.92 ± 0.62 \\
		BDLA\cite{zheng2023bdla} & Conv4-64 & 48.53 ± 0.87  & 70.07 ± 0.70  & \textbf{64.41 ± 0.84}  & 89.04 ± 0.45  &  50.59 ± 0.97  & 75.36 ± 0.72 \\
		\cmidrule{1-8}  
		DLDA($k=1$) &  Conv4-64 & \textbf{49.44 ± 0.85}  & 69.36 ± 0.69 & 60.86 ± 0.82 & 89.50 ± 0.41 & \textbf{55.12 ± 0.86} & 74.46 ± 0.65\\
		DLDA+Improved $k$-NN($k=3$) &  Conv4-64 & 49.08 ± 0.83 & \textbf{70.16 ± 0.67} & 60.04 ± 0.83 & \textbf{89.62 ± 0.42} & 54.53 ± 0.85 & \textbf{75.85 ± 0.68} \\
		\bottomrule
	\end{tabular*}
\end{sidewaystable}

\subsubsection{Results on the sensitivity to $k$}\label{subsubsec9}
\begin{figure}[H]
	\centering
	\begin{minipage}[t]{0.48\textwidth}
		\centering
		\includegraphics[width=6.5cm]{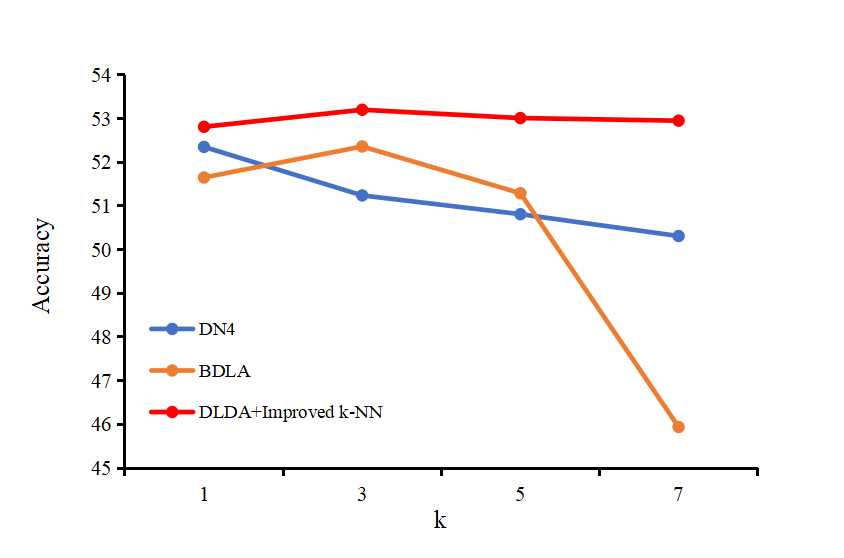}
		\caption{5-way 1-shot classification on the MiniImageNet dataset}\label{fig5}
	\end{minipage}
	\begin{minipage}[t]{0.48\textwidth}
		\centering
		\includegraphics[width=6.5cm]{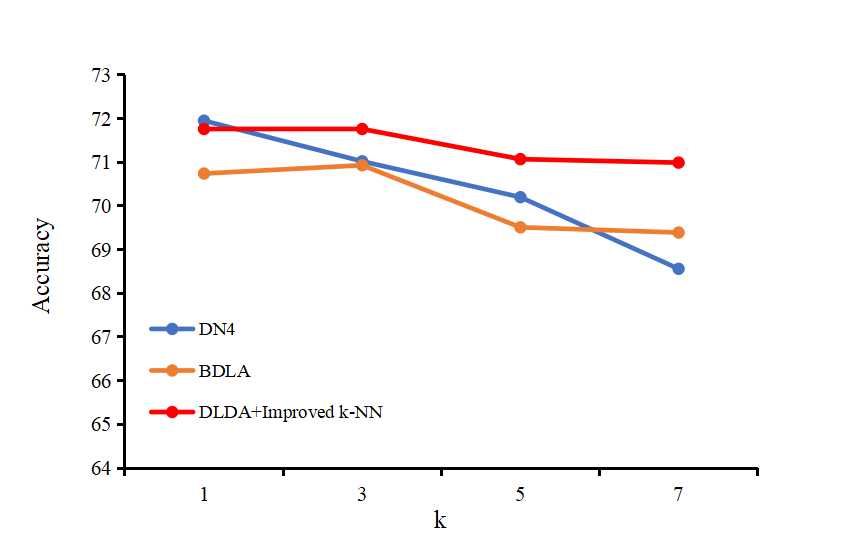}
		\caption{5-way 5-shot classification on the MiniImageNet dataset}\label{fig6}
	\end{minipage}
\end{figure}

Observe from the previous studies on $k$-NN involved few-shot learning methods, their numerical performance is usually dependent on the choice of $k$, which requires a large number of experiments such that it is set appropriately. 
Specifically, from Table \ref{tab3} we can find that in the case of 5-way 1-shot, the accuracy of the DN4 model drops from $52.35\%$ at $k=1$ to $50.31\%$ at $k=7$, with a fluctuation of $2.04\%$, and the one of the BDLA model reaches its peak $52.36\%$ at $k=3$ but slides to $45.94\%$ at $k=7$, with a fluctuation of $6.42\%$,
while the fluctuation of our proposed DLDA model added with an improved $k$-NN is a much smaller $0.39\%$. 
To address this issue, we provide a visual relationship between the selection of $k$ and the corresponding accuracy in Fig. \ref{fig5} and Fig. \ref{fig6}, which shows that our method is apparently less sensitive to the choices of $k$ over DN4 and BDLA, both in the cases of 5-way 1-shot and 5-way 5-shot.

\begin{table}[h]
	\caption{Average accuracy with different $k\ (k=1,3,5,7)$ on the MiniImageNet dataset}\label{tab3}
	\begin{tabular*}{\textwidth}{@{\extracolsep\fill}lccccc}
		\toprule%
		\textbf{Model}  & \multicolumn{5}{c}{\textbf{5-Way Accuracy (\%)} } \\
		\cmidrule{2-6}  
		&  &  $k=1$ & $k=3$ & $k=5$ & $k=7$ \\
		\midrule
		DN4\cite{li2019revisiting} & 1-shot & 52.35 & 51.24 & 50.81 & 50.31 \\
		\cmidrule{2-6}  
		& 5-shot & 71.95 & 71.02 & 70.2 & 68.56 \\
		\cmidrule{1-6}  
		BDLA\cite{zheng2023bdla} & 1-shot & 51.65 & 52.36 & 51.29 & 45.94 \\
		\cmidrule{2-6}  
		& 5-shot & 70.74 & 70.93 & 69.51 & 69.39 \\
		\cmidrule{1-6}  
		DLDA+Improved $k$-NN & 1-shot & 52.81 & 53.20 & 53.01 & 52.95 \\
		\cmidrule{2-6}  
		& 5-shot & 71.76 & 71.76 & 71.07 & 70.99 \\
		\bottomrule
	\end{tabular*}
\end{table}

\subsubsection{Results on cross-domain classification} \label{subsubsec10}
The cross-domain classification is to 
generalize the model that is pre-trained on the source domain to the target domain,
which appears to be an important indicator of measuring the ability of handling the differences among multiple sources in few-shot learning.
In our experiments, considering that the types of the images in the MiniImageNet dataset differ significantly from the ones in the three fine-grained datasets, we use the former dataset as the source domain to train the model and use the latter three datasets as the target domain to test the model. Moreover, we select three models, Prototypical Net \cite{snell2017prototypical}, DN4 \cite{li2019revisiting} and BDLA \cite{zheng2023bdla} for comparison. 
It can be obviously seen from Table \ref{tab4} that our method is the best performer under the domain shift.
This result indicates that aside from an improved classification performance and higher robustness to the parameter $k$, our method has a better cross-domain generalization capability over other competitors, which may be because the DLDA model together with the modified $k$-NN makes the local descriptors extracted by the embedded network more discriminative and transferable.

\begin{table}[h]
	\caption{Average accuracy with $95\%$ confidence intervals on the fine-grained datasets using the model trained on the MiniImageNet dataset}\label{tab4}
	\begin{tabular*}{\textwidth}{@{\extracolsep\fill}lccccc}
		\toprule%
		\textbf{Dataset}  &  & Prototypical Net & DN4 & BDLA & DLDA+ \\
		&  &  & & & Improved $k$-NN \\
		\midrule
		Stanford Dogs & 1-shot & 33.11 ± 0.64  & 36.32 ± 0.68 & 35.55 ± 0.66 & \textbf{37.10 ± 0.70} \\
		& 5-shot & 45.94 ± 0.65 & 53.43 ± 0.71 & 52.64 ± 0.69 & \textbf{53.99 ± 0.70} \\
		\cmidrule{1-6}  
		Stanford Cars & 1-shot & 29.10 ± 0.75 & 30.77 ± 0.57 & 30.62 ± 0.58 & \textbf{31.48 ± 0.56} \\
		& 5-shot & 38.12 ± 0.60 & 46.93 ± 0.62 & 45.99 ± 0.61 & \textbf{49.63 ± 0.66} \\
		\cmidrule{1-6}  
		CUB-200 & 1-shot & 39.39 ± 0.68 & 39.89 ± 0.73 & 40.40 ± 0.76 & \textbf{41.36 ± 0.74} \\
		& 5-shot & 56.06 ± 0.66 & 59.03 ± 0.71 & 58.23 ± 0.72 & \textbf{60.02 ± 0.71} \\
		\bottomrule
	\end{tabular*}
\end{table}

\section{Conclusions}\label{sec5}
In this paper, we focus on the few-shot image classification, and develop a new method for this problem, which includes a Discriminative Local Descriptors Attention (DLDA) model and an improved $k$-NN based classification model. 
Inspired by Fisher Score,
the DLDA model gives a weight to each local descriptor in the support set for highlighting the representative local features before the image-to-class classification. 
Based on the idea that the value of the information can be
partly reflected by the distance, in the final classification,
the improved $k$-NN model assigns larger weights to those local descriptors that are closer to the query point.
Extensive experimental results on the benchmark datasets illustrate that, compared with the
state-of-the-art few-shot learning methods,
the proposed method obtains a higher accuracy and a lower sensitivity to the parameter $k$, 
especially on the MiniImageNet dataset.
In addition, it also shows to be more capable of dealing with the situation of domain shift.

\bibliographystyle{siam}
\bibliography{sn-bibliography}

\begin{thebibliography}{10}

\bibitem{finn2017model}
{\sc C.~Finn, P.~Abbeel, and S.~Levine}, {\em Model-agnostic meta-learning for
  fast adaptation of deep networks},  (2017).

\bibitem{garcia2017few}
{\sc V.~Garcia and J.~Bruna}, {\em Few-shot learning with graph neural
  networks}, arXiv preprint arXiv:1711.04043,  (2017).

\bibitem{NIPS2014_5ca3e9b1}
{\sc I.~Goodfellow, J.~Pouget-Abadie, M.~Mirza, B.~Xu, D.~Warde-Farley,
  S.~Ozair, A.~Courville, and Y.~Bengio}, {\em Generative adversarial nets}, in
  Advances in Neural Information Processing Systems, Z.~Ghahramani, M.~Welling,
  C.~Cortes, N.~Lawrence, and K.~Weinberger, eds., vol.~27, 2014.

\bibitem{jamal2019task}
{\sc M.~A. Jamal and G.-J. Qi}, {\em Task agnostic meta-learning for few-shot
  learning}, in Proceedings of the IEEE/CVF Conference on Computer Vision and
  Pattern Recognition, 2019, pp.~11719--11727.

\bibitem{khosla2011novel}
{\sc A.~Khosla, N.~Jayadevaprakash, B.~Yao, and F.-F. Li}, {\em Novel dataset
  for fine-grained image categorization: Stanford dogs}, in Proc. CVPR workshop
  on fine-grained visual categorization (FGVC), vol.~2, Citeseer, 2011.

\bibitem{kingma2014adam}
{\sc D.~P. Kingma and J.~Ba}, {\em Adam: A method for stochastic optimization},
   (2014).

\bibitem{koch2015siamese}
{\sc G.~Koch, R.~Zemel, R.~Salakhutdinov, et~al.}, {\em Siamese neural networks
  for one-shot image recognition}, in ICML deep learning workshop, vol.~2,
  Lille, 2015.

\bibitem{krause20133d}
{\sc J.~Krause, M.~Stark, J.~Deng, and L.~Fei-Fei}, {\em 3d object
  representations for fine-grained categorization}, in Proceedings of the IEEE
  international conference on computer vision workshops, 2013, pp.~554--561.

\bibitem{lake2011one}
{\sc B.~Lake, R.~Salakhutdinov, J.~Gross, and J.~Tenenbaum}, {\em One shot
  learning of simple visual concepts}, in Proceedings of the annual meeting of
  the cognitive science society, vol.~33, 2011.

\bibitem{li2020more}
{\sc H.~Li, L.~Yang, and F.~Gao}, {\em More attentional local descriptors for
  few-shot learning}, in Artificial Neural Networks and Machine Learning --
  ICANN 2020, Cham, 2020, Springer International Publishing, pp.~419--430.

\bibitem{li2017feature}
{\sc J.~Li, K.~Cheng, S.~Wang, F.~Morstatter, R.~P. Trevino, J.~Tang, and
  H.~Liu}, {\em Feature selection: A data perspective}, ACM computing surveys
  (CSUR), 50 (2017), pp.~1--45.

\bibitem{li2019revisiting}
{\sc W.~Li, L.~Wang, J.~Xu, J.~Huo, Y.~Gao, and J.~Luo}, {\em Revisiting local
  descriptor based image-to-class measure for few-shot learning}, in
  Proceedings of the IEEE/CVF Conference on Computer Vision and Pattern
  Recognition, 2019, pp.~7260--7268.

\bibitem{liu2018learning}
{\sc Y.~Liu, J.~Lee, M.~Park, S.~Kim, and Y.~Yang}, {\em Transductive
  propagation network for few-shot learning},  (2018).

\bibitem{mehrotra2017generative}
{\sc A.~Mehrotra and A.~Dukkipati}, {\em Generative adversarial residual
  pairwise networks for one shot learning},  (2017).

\bibitem{mishra2017simple}
{\sc N.~Mishra, M.~Rohaninejad, X.~Chen, and P.~Abbeel}, {\em Meta-learning
  with temporal convolutions},  (2017).

\bibitem{paszke2019pytorch}
{\sc A.~Paszke, S.~Gross, F.~Massa, A.~Lerer, J.~Bradbury, G.~Chanan,
  T.~Killeen, Z.~Lin, N.~Gimelshein, L.~Antiga, et~al.}, {\em Pytorch: An
  imperative style, high-performance deep learning library}, Advances in neural
  information processing systems, 32 (2019).

\bibitem{qi2018low}
{\sc H.~Qi, M.~Brown, and D.~G. Lowe}, {\em Low-shot learning with imprinted
  weights}, in Proceedings of the IEEE conference on computer vision and
  pattern recognition, 2018, pp.~5822--5830.

\bibitem{ravi2017optimization}
{\sc S.~Ravi and H.~Larochelle}, {\em Optimization as a model for few-shot
  learning}, in International conference on learning representations, 2017.

\bibitem{santoro2016meta}
{\sc A.~Santoro, S.~Bartunov, M.~Botvinick, D.~Wierstra, and T.~Lillicrap},
  {\em Meta-learning with memory-augmented neural networks}, in International
  conference on machine learning, 2016, pp.~1842--1850.

\bibitem{schaul2010metalearning}
{\sc T.~Schaul and J.~Schmidhuber}, {\em Metalearning}, Scholarpedia, 5 (2010),
  p.~4650.

\bibitem{schwartz2018delta}
{\sc E.~Schwartz, L.~Karlinsky, J.~Shtok, S.~Harary, M.~Marder, A.~Kumar,
  R.~Feris, R.~Giryes, and A.~Bronstein}, {\em Delta-encoder: an effective
  sample synthesis method for few-shot object recognition}, Advances in neural
  information processing systems, 31 (2018).

\bibitem{snell2017prototypical}
{\sc J.~Snell, K.~Swersky, and R.~Zemel}, {\em Prototypical networks for
  few-shot learning}, Advances in neural information processing systems, 30
  (2017).

\bibitem{sung2018learning}
{\sc F.~Sung, Y.~Yang, L.~Zhang, T.~Xiang, P.~H. Torr, and T.~M. Hospedales},
  {\em Learning to compare: Relation network for few-shot learning}, in
  Proceedings of the IEEE conference on computer vision and pattern
  recognition, 2018, pp.~1199--1208.

\bibitem{takahashi2018ricap}
{\sc R.~Takahashi, T.~Matsubara, and K.~Uehara}, {\em Ricap: Random image
  cropping and patching data augmentation for deep cnns}, in Asian conference
  on machine learning, PMLR, 2018, pp.~786--798.

\bibitem{thrun2012learning}
{\sc S.~Thrun and L.~Pratt}, {\em Learning to learn}, Springer, New {Y}ork,
  2012.

\bibitem{vinyals2016matching}
{\sc O.~Vinyals, C.~Blundell, T.~Lillicrap, D.~Wierstra, et~al.}, {\em Matching
  networks for one shot learning}, Advances in neural information processing
  systems, 29 (2016).

\bibitem{welinder2010caltech}
{\sc P.~Welinder, S.~Branson, T.~Mita, C.~Wah, F.~Schroff, S.~Belongie, and
  P.~Perona}, {\em Caltech-ucsd birds 200},  (2010).

\bibitem{zeng2009pseudo}
{\sc Y.~Zeng, Y.~Yang, and L.~Zhao}, {\em Pseudo nearest neighbor rule for
  pattern classification}, Expert Systems with Applications, 36 (2009),
  pp.~3587--3595.

\bibitem{zhang2020deepemd}
{\sc C.~Zhang, Y.~Cai, G.~Lin, and C.~Shen}, {\em Deepemd: Few-shot image
  classification with differentiable earth mover's distance and structured
  classifiers}, in Proceedings of the IEEE/CVF conference on computer vision
  and pattern recognition, 2020, pp.~12203--12213.

\bibitem{zhang2017mixup}
{\sc H.~Zhang, M.~Ciss{\'{e}}, Y.~N. Dauphin, and D.~Lopez{-}Paz}, {\em mixup:
  Beyond empirical risk minimization},  (2017).

\bibitem{zhang2018metagan}
{\sc R.~Zhang, T.~Che, Z.~Ghahramani, Y.~Bengio, and Y.~Song}, {\em Metagan: An
  adversarial approach to few-shot learning}, Advances in neural information
  processing systems, 31 (2018).

\bibitem{zheng2023bdla}
{\sc Z.~Zheng, X.~Feng, H.~Yu, X.~Li, and M.~Gao}, {\em Bdla: Bi-directional
  local alignment for few-shot learning}, Applied Intelligence, 53 (2023),
  pp.~769--785.

\end{thebibliography}

\addcontentsline{toc}{section}{References}

\end{document}